\documentclass{article}
\usepackage{kr}

\usepackage{times}
\usepackage{soul}
\usepackage{url}
\usepackage[hidelinks]{hyperref}
\usepackage[utf8]{inputenc}
\usepackage[small]{caption}
\usepackage{graphicx}
\usepackage{amsmath}
\usepackage{amsthm}
\usepackage{booktabs}
\usepackage{algorithm}
\usepackage{algorithmic}
\newtheorem{example}{Example}

\title{Data Augmentation Techniques for Process Extraction from Scientific Publications}

\author{%
Yuni Susanti
\affiliations
Artificial Intelligence Lab., Fujitsu Limited, Japan\\
\emails
susanti.yuni@fujitsu.com
}

\begin{document}

\maketitle



\begin{abstract}
We present data augmentation techniques for process extraction tasks in scientific publications. We cast the process extraction task as a sequence labeling task where we identify all the entities in a sentence and label them according to their process-specific roles. The proposed method attempts to create meaningful augmented sentences by utilizing (1) process-specific information from the original sentence, (2) role label similarity, and (3) sentence similarity. We demonstrate that the proposed methods substantially improve the performance of the process extraction model trained on chemistry domain datasets, up to 12.3 points improvement in performance accuracy (F-score). The proposed methods could potentially reduce overfitting as well, especially when training on small datasets or in a low-resource setting such as in chemistry and other scientific domains.

\end{abstract}

\section{Introduction}
\label{sec:intro}
In the scientific field, scientists often refer to scientific publications for conducting their research and observation. For example, in the material science and chemistry field, the process of creating a new material involves scientists seeking information from scientific publications, e.g. to understand the relationship between material structure and properties. This is a common practice since experimental observations have been the primary means to obtain various properties of materials. Other important information from experimental observations includes the step-by-step process of conducting the experiment itself. Scientists could refer to that information to replicate, or conduct similar experiments, or in material science, as a reference when creating new materials. Thus, access to structured information on experimental processes is needed to improve the efficiency of their work.

The process extraction task aims to automatically extract process-specific information from unstructured text sources, e.g. process sentences in the experimental section of scientific publications. This task falls under the Natural Language Processing (NLP) task. This task suffers from an insufficient number of labeled resources, and data annotation is difficult for domain-specific tasks. One of the techniques for increasing the size of labeled data is using \textbf{data augmentation}, which refers to strategies for increasing the diversity of training examples without explicitly collecting new data~\cite{feng-etal-2021-survey}. In this study, we explore data augmentation techniques for process extraction tasks from scientific publications such as in chemistry domain.

Previous works have proposed data augmentation techniques for various NLP tasks\cite{wang-yang-2015-thats,DBLP:journals/corr/XieWLLNJN17,kobayashi-2018-contextual,DBLP:journals/corr/abs-1804-09541,chen-etal-2021-hiddencut,wei-zou-2019-eda}. For instance, in the text classification task, \cite{wang-yang-2015-thats} proposed the use of neighboring words in continuous representations to create new instances to augment the training data, while \cite{kobayashi-2018-contextual} used predictive language models for synonym replacement. A study by~\cite{DBLP:journals/corr/abs-1804-09541} generated new data by backtranslation, while~\cite{DBLP:journals/corr/XieWLLNJN17} used data noising as a smoothing method.~\cite{chen-etal-2021-hiddencut} proposed HiddenCut, which is able to learn more generalizable features in natural language generation tasks. 

Easy Data Augmentation, or EDA~\cite{wei-zou-2019-eda}, comprehensively explored text editing-based data augmentation techniques for text classification: synonym replacement, random swap, random insertion, and random deletion. The results show a strong performance gain despite the simplicity of the techniques. Inspired by their work, we explore data augmentation techniques for process extraction from scientific publications. We also explore the possibility of creating meaningful new data from the existing data by preserving the meaning of the original information of the existing data. 

To summarize, our contributions are:
\begin{enumerate}
    \item Introducing new data augmentation techniques for process extraction tasks utilizing process-specific information from the original sentence. By this method, we aim to preserve the original meaning of the original data (i.e. original sentence) in the augmented sentence results.
    \item Demonstrating the effectiveness of the proposed methods through extensive evaluation experiments. The proposed method outperforms the original model by a wide margin, up to 12.3 points F-score improvement in the performance. 
\end{enumerate}

\section{Process Extraction Task}
\label{sec:proext}
Process extraction task aims to extract entities\footnote{In this study, we broaden the definition of ``entity'' to refer to any word/token deemed important in a process sentence.} from a process sentence and label them with process-specific entity-type labels such as operation/process, material, and condition. These labels define the roles of the entities in the process sentences. In this paper, we focus only on process sentences, which we define as sentences containing \textit{process predicates}. Process sentences are commonly found in scientific papers used for describing the experiment process. The following shows an example of a process sentence, consisting of one process predicate \textbf{biotinylated}.
\begin{example}
\label{ex1}
    Antibodies were  \underline{biotinylated} using a 5-fold molar excess of biotin-LC-NHS ester.
\end{example}

In this study, \textit{process predicate} is defined as an expression representing the operation of a process, e.g. the word \textit{biotinylated} in the Example~\ref{ex1}. In this work, we assume that a process sentence contains one or more process predicates. Given a chunked text as the input, the process extraction task aims to identify all the entities\footnote{Entity span, precisely, since an entity can be represented as a word or sequence of words} including their types/ roles in a process sentence. We cast this problem as a sequence labeling task.
\section{Proposed Approach}
\label{sec:dataaug}
To our knowledge, only a few studies have been conducted on data augmentation specifically addressing sequence labeling tasks. SeqVat~\cite{chen-etal-2020-seqvat} applies Virtual Adversarial Training (VAT) to sequence labeling models with Conditional Random Field (CRF), while \cite{andreas-2020-good} proposed a method aimed at providing a compositional inductive bias in sequence models by creating synthetic training examples. Similar to EDA~\cite{wei-zou-2019-eda}, Snippext~\cite{Miao10.1145/3366423.3380144} also employs text-editing based methods for data augmentation such as token/span replacement, insertion, and deletion, and further fine-tunes a language model through semi-supervised learning with the augmented data. Text-editing techniques such as EDA and Snippext can be easily applied to our target task; however, it showed performance decreases in our experiment. Considering the different nature of the tasks, this result is expected. In the sequence labeling task, swapping the words or randomly inserting and deleting the entities could be the main reason for the decrease in performance since those techniques change the grammatical structure of the sequence. 

In this work, we introduce data augmentation techniques for process extraction tasks. The idea is to replace an entity with another entity having the same entity type in order to create a new sentence. However, in the methods that we propose, we substitute all entities in one sentence with entities taken from other sentences with some sort of similarity to the original sentence. Therefore, we focus on the method of selecting a sentence to serve as the source of entity replacement.
Similar to EDA~\cite{wei-zou-2019-eda}, our method does not require domain-specific dictionary or any dictionary at all.\footnote{Nonetheless, EDA~\cite{wei-zou-2019-eda} uses WordNet lexical dictionary to find synonyms in the synonym replacement method}

In the proposed method, we suppose a sentence $S$ has a set of entities $e$ and a pattern $pat$, formally $S$ = $(S_{e}$, $S_{pat})$. The \textit{pattern} here refers to the grammatical structure of the sentence including the order of words, stop word choices, etc. For a set of entities in an \textbf{input sentence} ($input_{e}$), a new sentence using the entity set is created using a sentence pattern from a \textbf{source sentence} ($source_{pat}$). This new sentence, or the \textbf{augmented sentence}, would have the entities from the input sentence and the pattern from the source sentence ($augmented = input_{e}, source_{pat}$). Thus, given an input sentence, the proposed data augmentation method is twofold process:
\begin{enumerate}
    \item \textbf{Source sentence selection}: Selecting a source sentence to be the pattern for generating the new sentence. The candidates for the source sentences are all sentences in the training data except for the input sentence. To make $k$-new sentences, $k$-number of source sentences are selected.
    \item \textbf{Entity replacement}: An augmented sentence is created by substituting the entities in the source sentence having the same entity types as the corresponding entities in the input sentence. If there is more than one entity with the same type, we employ entity vector similarity methods to choose the appropriate replacement.
\end{enumerate}

In the following, we describe the proposed methods for source sentence selection in detail. 

\subsection{Label similarity-based method (LSIM)} 
\label{sec:lsim}
In this method, we simply choose the sentence with the biggest number of label overlaps with the input sentence as the source sentence. The idea is that sentences with a high label overlap should be similar in structure, as well. For instance, the following input and source sentence pairs have 100\% label overlaps: 

\begin{itemize}
    \item [] \textbf{INPUT: Oxalic$_{MAT}$ acid$_{MAT}$ were$_{O}$ dissolved$_{PP}$ in$_{O}$ deionized$_{DESC}$ water$_{MAT}$}
    \item [] \textbf{SOURCE: Borac$_{MAT}$ acid$_{MAT}$ was$_{O}$ added$_{PP}$ to$_{O}$ boiling$_{DESC}$ alcohol$_{MAT}$}
\end{itemize}

Both sentences have the same sentence structure; thus, it is safe to use the structure of the source sentence together with the set of entities from the input sentence, and vice versa. We further rank the source sentence candidates according to their label overlaps and take $k$-highest scoring sentences as the source sentences to create $k$-new augmented sentences. For the above input-source sentence pairs, the augmented sentence would become the following:

\begin{itemize}
    \item [] \textbf{Augmented}: Oxalic acid was added to deionized water
\end{itemize}

The augmented sentence uses the pattern from the source sentence and entities from the input sentence, e.g. \textit{oxalic acid} substitutes \textit{borac acid} since they have the same entity type, i.e. \textit{MAT} (material). As we can see, the created augmented sentence is structurally correct although the meaning slightly differs. 

\subsection{Process similarity-based method} The augmented sentences generated by the LSIM method often loses the meaning of the original sentence. It might also form into a non-sense sentence, albeit to be structurally correct. For instance, in the example of the LSIM method in Section~\ref{sec:lsim}, the ``oxalic acid" in the original sentence changes from ``\textit{were dissolved in deionized water}" to ``\textit{added in deionized water}" in the generated augmented sentence. This could be problematic since instead of providing useful information to the model, they can be noise to the training data. Especially in the chemistry domain, a change in some properties of a process sentence describing a specific chemical reaction, e.g. change of material/condition, could result in the said reaction not occurring correctly. To avoid this problem, we choose a sentence describing a similar process as the source sentence in this method.
Based on our observation, to preserve the meaning of the original process sentence, the most important part is the process predicate, as described in the example of \textit{dissolved} vs \textit{added} above. Therefore, we focus on the similarity between process predicates to find similar process sentences. In this work, we leverage the Word2vec pre-trained model~\cite{Kim2017} to calculate the vector similarity of the process predicate of the input sentence and the source sentence candidates. After filtering based on its label similarity, the sentence with the highest process predicate similarity score with those of the input sentence is chosen as the source sentence. Since one sentence could have more than one process predicate, we make combinations of process predicates from the sentence pair and calculate the similarity between them. We propose process predicate similarity calculation \textbf{PSIM} and \textbf{PSIM-A} as in the following:\\
\begin{equation}
    \text{PSIM} = \frac{1}{|I||S|} \sum_{i=1}^{|I|} \sum_{j=1}^{|S|}SIM(I_{i}, S_{j}) \\
\label{eq2}
\end{equation}

\begin{equation}
    \text{PSIM-A} = \frac{1}{|I|} \sum_{i=1}^{|I|} \max_{j=1,..,|S|}SIM(I_{i}, S_{j}) \\
\label{eq}
\end{equation}

where $I, S$ represent process predicates in input and source sentences, respectively, and $SIM(I,S)$ is the similarity score between them. PSIM simply averages the similarity scores of all process predicates, while PSIM-A tries to align the process predicate by taking the maximum score for each process predicate in the input sentence. The following shows an example of the input-source pair and the augmented sentence result using this method. 

\begin{enumerate}
\item [] \textbf{Input}: The pH of the \underline{mixed} solution was \underline{adjusted} to 6 with ammonia (28\%).
\item [] \textbf{Source}: The pH value of the K2HPO4 was \underline{adjusted} to 0.1 \underline{using} 0.1 M phospate solutions.
\item [] \textbf{Augmented}: The pH value of the solution was \underline{adjusted} to 6 \underline{using} 28\% ammonia.
\end{enumerate}

The result shows that the original meaning of the input sentence is preserved in the resulting augmented sentence to some extent. In addition, by keeping the original process predicates (of the source sentence), the augmented sentence maintains the structure of the source sentence. We keep the process predicate because we regard the pattern ($S_{pat}$) as a set with its predicate. 

\subsection{Sentence similarity-based method} A more straightforward method to find a similar process sentence is by using sentence similarity. We experiment with word vector sentence similarity by calculating the average vector for all words in input and source sentences and further applying cosine similarity between the vectors (\textbf{SSIM}). We also implement Word Mover`s Distance (\textbf{WMD}), which measures the dissimilarity between texts as the minimum amount of distance that the embedded words of one document need to ``travel`` in order to reach the embedded words of another document~\cite{wmd2015}.

As a comparison, the following methods serve as baselines in this work: 1) random-entity (\textbf{RE}): randomly substitutes entities with other entities in the data, 2) ranked-entity (\textbf{RAE}): an entity is replaced with the most similar entity in the data. We rank the entity based on its word vector similarity.

\section{Experiment Setup}
\label{exp}
To test the effectiveness of the data augmentation, we conducted experiments on the process extraction models trained with and without the augmented data generated with the data augmentation techniques described in the previous section. We used a dataset of 235 material synthesis procedures annotated by experts, comprising 1896 examples for training and 210 examples for testing.
The dataset is collected from paragraphs describing material synthesis in 2.5 million publications~\cite{InMat2020,mysore-etal-2019-materials}. There are in total 21 entity types including material, operation, condition, description, etc. We conducted the experiments using Bi-LSTM with CRF model, which shows good performance in sequence labeling tasks~\cite{DBLP:journals/corr/HuangXY15}. 

\captionsetup[table]{skip=1pt}
\begin{table*}
	\small
	\centering
	\begin{tabular}{cccccccccccc}
	\toprule
\multicolumn{3}{c}{training set size} & \multicolumn{1}{c}{\textit{k}} & \multicolumn{1}{c}{org model} & \multicolumn{2}{c}{baseline} & \multicolumn{1}{c}{LSIM} & \multicolumn{2}{c}{process-sim} & \multicolumn{2}{c}{sentence-sim} \\
fraction & original(org) & org+aug &  &  & RE & RAE &  & PSIM & PSIM-A & SSIM & WMD \\
\midrule
10\% & 189 & 1004 & 5 & 46.3 & 50.8 & 52.2 & 52.7 & 54.1 & 52.8 & \textbf{54.9} & 54.7 \\
20\% & 379 & 1989 & 5 & 53.2 & 54.7 & 58.8 & 56.0 & 58.7 & \textbf{59.0} & 57.3 & 58.6 \\
30\% & 568 & 2963 & 5 & 57.8 & 58.8 & 58.6 & 60.2 & 61.6 & 60.7 & \textbf{61.9} & 59.2 \\
40\% & 758 & 4063 & 5 & 62.2 & 63.3 & 63.6 & \textit{61.9} & 65.1 & \textbf{66.6} & 62.6 & \textit{61.6} \\
50\% & 947 & 5027 & 5 & 63.2 & 64.5 & 64.1 & \textit{61.1} & \textbf{65.5} & 64.9 & 64.1 & 63.8 \\
60\% & 1136 & 6021 & 5 & 64.5 & 64.7 & 67.7 & 66.8 & 66.4 & 68.1 & \textbf{68.5} & 65.8 \\
70\% & 1326 & 7056 & 5 & 66.2 & 67.5 & 67.3 & \textit{64.9} & \textbf{69.2} & 67.5 & 67.9 & 68.6 \\
80\% & 1517 & 8077 & 5 & 66.7 & 67.0 & \textit{65.5} & \textit{65.9} & 68.2 & 69.0 & \textbf{71.0} & 68.5 \\
90\% & 1706 & 9121 & 5 & 66.0 & 69.4 & 69.2 & 68.7 & 69.0 & 69.5 & 69.3 & \textbf{70.2} \\
100\% & 1896 & 10106 & 5 & 68.2 & 69.8 & 71.1 & 70.1 & \textbf{72.5} & 71.1 & 69.8 & 69.3\\
\multicolumn{3}{l}{average performance gains} & & & 1.62 & 2.38 & 1.40 & \textbf{3.60} & 3.49 & 3.30 & 2.60 \\
\midrule
10\% & 189 & 1493 & 8 & 46.3 & 53.2 & 51.0 & 53.4 & 51.4 & 53.7 & \textbf{56.2} & 52.9 \\
40\% & 758 & 6046 & 8 & 62.2 & 61.5 & 64.3 & 62.4 & 64.4 & \textbf{66.7} & 65.6 & 62.5 \\
90\% & 1706 & 13570 & 8 & 66.0 & 66.5 & 69.6 & 67.8 & 69.6 & 69.3 & 67.3 & \textbf{70.2} \\
\midrule
10\% & 189 & 2797 & 16 & 46.3 & 54.8 & 51.4 & 56.7 & \textbf{58.6} & 58.2 & 56.4 & 54.0 \\
40\% & 758 & 11334 & 16 & 62.2 & 63.3 & 63.7 & \textit{62.1} & 64.8 & 66.8 & \textbf{67.4} & 66.2 \\
90\% & 1706 & 25434 & 16 & 66.0 & 68.1 & 67.2 & \textit{65.9} & 70.2 & 69.5 & \textbf{70.3} & 68.9 \\
\bottomrule
\end{tabular}
\vspace{+2mm}
\caption{F1 scores of data augmentation methods on the process extraction model by varying the training set fractions and the number of augmented sentences per original sentence ($k$). ``org model" denotes the original model, which is the model trained without any additional augmented data. \textbf{Bold} indicates the best score for each training set fraction; \textit{italic} indicates decrease in performance over the original model.} \medskip
\label{tab:peresult}
\end{table*}

\section{Result and Discussion}
\label{sec:pilot_study}
We run the experiments on original data and original data combined with the additional augmented data for the following training set fractions (\%): {10, 20, 30, 40, 50, 60, 70, 80, 90, 100}. The same hyperparameter settings were used in all experiments. We generate $k: 5, 8, 16$-augmented sentences per each original sentence. Table~\ref{tab:peresult} shows the results (F1 scores, in \%) of the process extraction model.

To sum up the results, almost all of the models trained including additional data generated with the data augmentation techniques outperform the original model trained without data augmentation. The improvement varies depending on the size of the data used in the training, ranging from 0.1\% to as high as 12.3\%. The best F1 score without data augmentation, 68.2, was achieved using all training data, i.e. 100\% fraction. However, the proposed model SSIM, which was trained using data augmentation, surpassed this number while using only a 60\% fraction of the training data, which shows the effectiveness of the proposed data augmentation methods. 
Table~\ref{tab:peresult} shows that the data augmentation has the most significant improvement when training on the smaller dataset fractions (10-20\% fractions), with performance improvement up to 12.3\%, (PSIM, 10\% data, $k$=16). Nevertheless, the improvements are reasonable on the bigger training set fractions (90-100\%) experiments, up to 4.3\% improvement in F score. Since overfitting tends to be more severe when training on small datasets, this result is encouraging since it shows that the proposed data augmentation techniques could potentially reduce the risk of overfitting, especially on the smaller-sized training sets.

While the improvements across all methods are more or less the same, the proposed PSIM, PSIM-A and SSIM methods constantly contributed to performance gain in all training set fractions experiments. They also outperform the baselines RE and RAE in most experiments. On the other hand, LSIM shows lower F1 scores than the baseline RAE in most experiments. This proves that keeping the original meaning of the input sentence in the augmented sentence was beneficial. 

Although sentence similarity-based methods (SSIM, WMD) outperform the baselines and LSIM, the average performance gains are lower compared to process similarity-based methods, i.e. the PSIM and PSIM-A methods. One explanation is that the sentence similarity-based methods might end up with \textit{too similar} source sentence, resulting in less variance in the generated augmented sentence. Thus, it might inadvertently add noise to the training data. Nevertheless, this result is encouraging because simply using the process predicate actually gave a better result compared to a more complicated sentence similarity-based method.

\paragraph {Does $k$ affect performance?} We also conducted experiments where we varied the number of augmented sentences per each input sentence ($k$=5, 8, 16). To summarize, performance gains were indeed increasing with the increase of $k$, but only on the small dataset fractions (10\% fraction). For example, in the experiment using a 10\% fraction of the data, the SSIM model yields 8.6, 9.9, and 10.1\% improvements in F scores using $k$=5, 8, 16, respectively. On the other hand, the improvements were found to be not significantly different across different \textit{k} in the experiments using bigger data fractions. 
\paragraph {Adding a pre-trained model} EDA~\cite{wei-zou-2019-eda} stated that it might not give substantial performance gains when training using pre-trained models, thus we conducted experiments where we included a pre-trained Word2vec model trained on 640k+ materials science articles~\cite{Kim2017}. The result shows some improvements, up to 4.2\% improvements with the process similarity-based methods PSIM and PSIM-A, and up to 3.9\% improvement with the sentence similarity-based method, while a small decrease of performance is observed in the LSIM model.

\section{Conclusion}
\label{sec:conclusion}
Our proposed data augmentation techniques attempt to create meaningful new sentences by utilizing process information (\textit{process predicate}) in a process sentence. We demonstrate that despite its simplicity, it substantially improves the performance of the process extraction model, up to 12.3\% F1 score improvement, and potentially reduces overfitting especially in low-resource settings such as when there are not enough datasets for training. The future work includes:
\begin{enumerate}
    \item Evaluating the proposed method for general sequence labeling tasks. The proposed methods are designed for process extraction task, however, we believe that the methods could be applied to general sequence labeling tasks.
    \item Evaluation on datasets of different domains. In this work, we report the evaluation results on material and chemistry scientific publications. We plan to conduct an evaluation on different domain as well.  
\end{enumerate}

\bibliographystyle{kr}
\bibliography{kr-sample}

\end{document}